\documentclass[10pt,twocolumn,letterpaper]{article}

\usepackage{caption}
\usepackage{diagbox}

\usepackage{times}
\usepackage{epsfig}
\usepackage{graphicx}
\usepackage{float}
\usepackage{wrapfig}
\usepackage{amsmath,amssymb,amsthm}
\usepackage{algorithm,algorithmicx,algpseudocode}
\usepackage{bm,xspace}
\usepackage{comment}
\usepackage{verbatim}
\usepackage{multirow}
\usepackage{balance}
\usepackage[hyphens]{url}
\usepackage{booktabs}
\usepackage{etoolbox,siunitx}
\usepackage{calc}
\usepackage{pifont,hologo}
\usepackage[usenames, dvipsnames]{color}
\usepackage{epigraph}
\usepackage{enumitem}
\usepackage{soul}
\usepackage{mdframed}
\usepackage{microtype}

\usepackage[super]{nth}
\usepackage{nicefrac}
\robustify\bfseries

\def\cc{\mathbf{c}}
\def\dd{\mathbf{d}}
\def\ee{\mathbf{e}}
\def\ff{\mathbf{f}}

\def\oo{\mathbf{o}}

\def\xx{\mathbf{x}}

\def\CC{\mathbf{C}}

\def\II{\mathbf{I}}

\def\MM{\mathbf{M}}

\def\PP{\mathbf{P}}

\def\SS{\mathbf{S}}

\def\XX{\mathbf{X}}

\def\xX{\mathcal{X}}

\def\btheta{{\bm\theta}}

\DeclareMathSymbol{@}{\mathord}{letters}{"3B}

\newcommand\mypara[1]{\vspace{1mm}\noindent\textbf{#1}}

\def\latex/{\LaTeX}
\def\bibtex/{\hologo{BibTeX}}

\usepackage{iccv}
\usepackage{times}
\usepackage{epsfig}
\usepackage{graphicx}
\usepackage{amsmath}
\usepackage{amssymb}
\usepackage{tikz} 
\usepackage{multirow}
\usepackage{subcaption}

\usepackage[breaklinks=true,bookmarks=false]{hyperref}

\usepackage[capitalize]{cleveref}
\crefname{section}{Sec.}{Secs.}
\Crefname{section}{Section}{Sections}
\Crefname{table}{Table}{Tables}
\crefname{table}{Tab.}{Tabs.}
\def\SS{\mathbf{S}}

\iccvfinalcopy 

\def\iccvPaperID{2987} 

\ificcvfinal\fi

\begin{document}

\title{CLNeRF: Continual Learning Meets NeRF}

\author{Zhipeng Cai$^*$\\
Intel Labs\\
{\small zhipeng.cai@intel.com}
\and
Matthias M\"uller\\
Intel Labs\\
{\small matthias.mueller.2@kaust.edu.sa}
}

\twocolumn[{%
\renewcommand\twocolumn[1][]{#1}%
\maketitle

 \begin{center}
    \centering
        \captionsetup{type=figure}
        \vspace{-2em}
 	\includegraphics[width=1\textwidth]{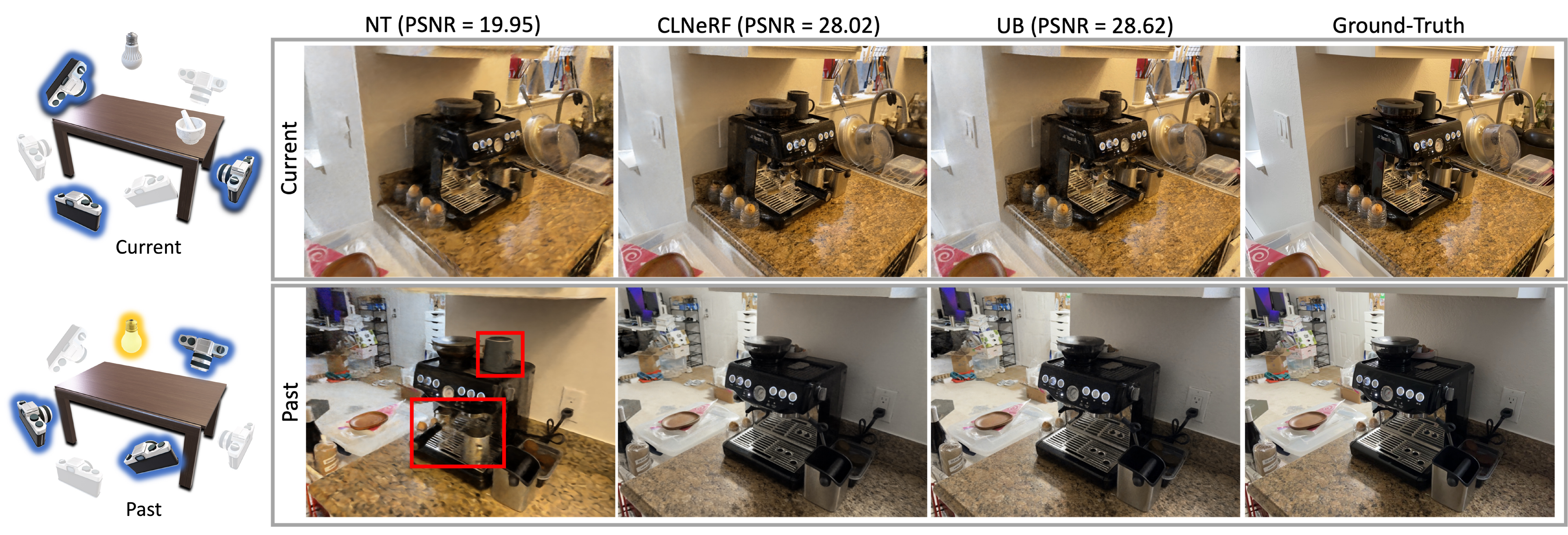}
 	\captionof{figure}{\textbf{Teaser.} This work studies continual learning for NeRFs. We propose a new benchmark -- \emph{World Across Time (WAT)} -- to study practical scenarios where images of a scene arrive as a sequence of multiple scans with appearance and geometry changes, over an extended period of time. We also propose an effective system -- \emph{CLNeRF} -- that can sequentially learn from these scans, without requiring stored historical images. The top and bottom rows show rendered novel views of the same scene for the current and past scans respectively. CLNeRF accurately renders both the current and the past scans, performing on-par with the upper bound model (UB) trained on all scans at once. Naively training (NT) on the sequence of scans overfits to the current scan, resulting in erroneous appearance (lightning) and geometry (extra cups marked by bounding boxes, which only exist in the current scan) for the past scan.}\label{fig:demo}
\end{center}
}]

\ificcvfinal\thispagestyle{empty}\fi

\begin{abstract}
Novel view synthesis aims to render unseen views given a set of calibrated images. In practical applications, the coverage, appearance or geometry of the scene may change over time, with new images continuously being captured. Efficiently incorporating such continuous change is an open challenge. Standard NeRF benchmarks only involve scene coverage expansion. To study other practical scene changes, we propose a new dataset, World Across Time (WAT), consisting of scenes that change in appearance and geometry over time. We also propose a simple yet effective method, CLNeRF, which introduces continual learning (CL) to Neural Radiance Fields (NeRFs). CLNeRF combines generative replay and the Instant Neural Graphics Primitives (NGP) architecture to effectively prevent catastrophic forgetting and efficiently update the model when new data arrives. We also add trainable appearance and geometry embeddings to NGP, allowing a single compact model to handle complex scene changes. Without the need to store historical images, CLNeRF trained sequentially over multiple scans of a changing scene performs on-par with the upper bound model trained on all scans at once. Compared to other CL baselines CLNeRF performs much better across standard benchmarks and WAT. The source code, a demo, and the WAT dataset are available at \url{https://github.com/IntelLabs/CLNeRF}.
\end{abstract}

\section{Introduction} \label{sec:intro}


Neural Radiance Fields (NeRFs) have emerged as the preeminent method for novel view synthesis. Given images of a scene from multiple views, NeRFs can effectively interpolate between them. However, in practical applications (\eg, city rendering~\cite{blocknerf}), the scene may change over time, resulting in a gradually revealed sequence of images with new scene coverage (new city blocks), appearance (lighting or weather) and geometry (new construction). Learning continually from such sequential data is an important problem.

Naive model re-training on all revealed data is expensive, millions of images may need to be stored for large scale systems~\cite{blocknerf}. Meanwhile, updating the model only on new data leads to catastrophic forgetting~\cite{catastrophic_forgetting}, \ie, old scene geometry and appearances can no longer be recovered (see \cref{fig:demo}). Inspired by the continual learning literature for image classification~\cite{cl_survey}, this work studies continual learning in the context of NeRFs to design a system that can learn from a sequence of scene scans without forgetting while requiring minimal storage.

Replay is one of the most effective continual learning algorithms; it trains models on a blend of new and historical data. \emph{Experience replay}~\cite{er} explicitly stores a tiny portion of the historical data for replay, while \emph{generative replay}~\cite{generative_replay} synthesizes replay data using a generative model (e.g., a GAN~\cite{gan}) trained on historical data. Experience replay is more widely used in image classification, since generative models are often hard to train, perform poorly on high resolution images, and introduce new model parameters. In contrast, NeRFs excel at generating high-resolution images, making them ideal candidates for generative replay. 

Motivated by this synergy between advanced NeRF models and generative replay, we propose \emph{CLNeRF}
which combines generative replay with Instant Neural Graphics Primitives (NGP)~\cite{ngp} to enable efficient model updates and to prevent forgetting \emph{without the need to store historical images}. CLNeRF also introduces trainable appearance and geometry embeddings into NGP so that various scene changes can be handled by a single model. Unlike classification-based continual learning methods whose performance gap to the upper bound model is still non-negligible~\cite{generative_replay}, the synergy between continual learning and advanced NeRF architectures allows CLNeRF to achieve a similar rendering quality as the upper bound model (see \cref{fig:demo}).



\mypara{Contributions:} (1) We study the problem of continual learning in the context of NeRFs. We present \emph{World Across Time (WAT)}, a practical continual learning dataset for NeRFs that contains scenes with real-world appearance and geometry changes over time. (2) We propose \emph{CLNeRF}, a simple yet effective continual learning system for NeRFs with minimal storage and memory requirements. Extensive experiments demonstrate the superiority of CLNeRF over other continual learning approaches on standard NeRF datasets and WAT.

\section{Related Work}\label{sec:related}

\mypara{NeRF.} Learning Neural Radiance Fields (NeRFs) is arguably the most popular technique for novel view synthesis (see~\cite{nerf_survey} for a detailed survey). Vanilla NeRF~\cite{nerf} represents a scene implicitly using neural networks, specifically, multi-layer perceptrons (MLPs). These MLPs map a 3D location and a view direction to their corresponding color and opacity. An image of the scene is synthesized by casting camera rays into 3D space and performing volume rendering. Though effective at interpolating novel views, vanilla NeRF has several limitations, for example, the slow training/inference speed. This problem is addressed by using explicit scene representations~\cite{sun2022direct, ngp}, or spatially-distributed small MLPs~\cite{kilonerf}. CLNeRF applies these advanced architectures to ensure efficient model updates during continual learning. Vanilla NeRF only considers static scenes; to handle varied lightning or weather conditions, trainable appearance embeddings are introduced~\cite{nerfw, blocknerf}. Transient objects in in-the-wild photos are handled by either introducing a transient MLP~\cite{nerfw} or using segmentation masks~\cite{blocknerf}. CLNeRF adopts these techniques to allow a single model to handle complex scene changes. Concurrent to this work, Chung et al.~\cite{meilnerf} also study NeRFs in the context of continual learning. However, they only consider static scenes and the vanilla NeRF architecture. We consider scenes with changing appearance/geometry, and introduce a new dataset to study such scenarios. The proper combination of continual learning and more advanced architectures also makes CLNeRF simpler (no extra hyperparameters) and much more effective at mitigating forgetting.

\mypara{Continual Learning.} 
Continual learning aims to learn from a sequence of data with distribution shifts, without storing historical data (see~\cite{cl_survey} for a detailed survey). Naive training over non-IID data sequences suffers from catastrophic forgetting~\cite{ewc} and performs poorly on historical data. A popular line of work regularizes the training objective to prevent forgetting~\cite{ewc, lwf}. However, since the regularization does not rely on historical data it is less effective in practice. Parameter isolation methods prevent forgetting by freezing (a subset of) neurons from previous tasks and use new neurons to learn later tasks~\cite{packnet, HAT}. Though remembrance can be guaranteed~\cite{packnet}, these methods have a limited capacity or grow the network significantly given a large number of tasks. Replay-based approaches use historical data to prevent forgetting. This historical data is either stored in a small replay buffer~\cite{er, gem}, or synthesized by a generative model~\cite{generative_replay}. 
Generative replay~\cite{generative_replay}, i.e., synthesizing historical data, is less effective for image classification, since the generative model introduces extra parameters, and performs poorly on high resolution images. In contrast, this work shows that advanced NeRF models and generative replay benefit from each other, since high quality replay data can be rendered without introducing new model parameters.

\section{Method}\label{sec:method}
\begin{figure*}[t]
	\center
 	\includegraphics[width=1\textwidth]{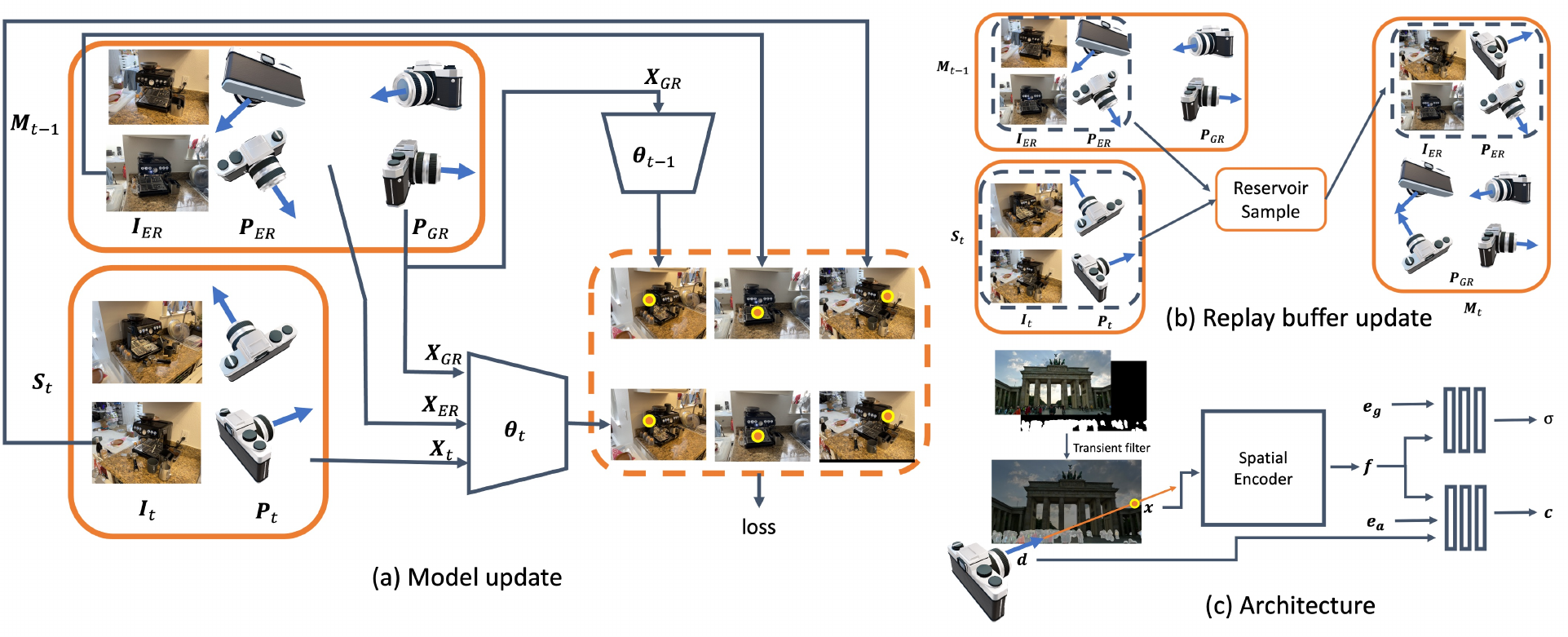}
	\caption{\textbf{System overview.} (a): At each time step $t$ of continual NeRF, a new set of data $\SS_t$ is generated. To update the model $\btheta_t$, we randomly generate in each training iteration a set of camera rays from camera parameters stored for experience replay ($\PP_\text{ER}$), generative replay ($\PP_\text{GR}$) and in the new data ($\PP_t$). For rays from new data ($\xX_t$) or experience replay ($\xX_\text{ER}$), the corresponding image color is used for supervision. For rays from generative replay, \ie, $\xX_\text{GR}$, we use the latest deployed model $\btheta_{t-1}$ to generate pseudo-labels for supervision. After training $\btheta_t$, we replace the previously deployed model $\btheta_{t-1}$ with $\btheta_t$, and update the replay buffer $\MM_{t}$. (b): To update the replay buffer $\MM_t$ under optional experience replay, we perform reservoir sampling over image-camera-parameter pairs in $\MM_{t-1}$ and $\SS_t$, and add into $\PP_\text{GR}$ camera parameters for all images not selected by reservoir sampling. (c) We use segmentation masks to filter transient objects, and apply appearance embeddings $\ee_a$ and geometry embeddings $\ee_g$ to the base architecture to handle scene changes at different time steps.}\label{fig:overview}
\end{figure*}

\subsection{Preliminaries}\label{sec:prelim}

Before introducing CLNeRF, we first review the basics of NeRFs, and formulate the problem of continual learning. 

\mypara{NeRF.} Given a set of images, NeRFs train a model parameterized by $\btheta$ that maps a 3D location $\xx\in\mathbb{R}^3$ and a view direction $\dd\in\mathcal{S}^3$ (a unit vector from the camera center to $\xx$) to the corresponding color $c(\xx, \dd| \btheta)\in[0,1]^3$ and opacity $\sigma(\xx| \btheta) \in [0,1]$. Given a target image view, we render the color for each pixel independently. For each pixel, we cast a ray from the camera center $\oo\in\mathbb{R}^3$ towards the pixel center, and sample a set of 3D points $\XX = \{\xx_i|\xx_i = \oo + \tau_i \dd\}$ along the ray, where $\tau_i$ is the euclidean distance from the camera center to the sampled point. Then, we render the color $\hat{\CC}(\XX)$ of the ray following the volume rendering~\cite{volume_render_survey} equation:
\begin{align}\label{eq:volume_render}
  \hat{\CC}(\XX) = \sum_{i} w_ic(\xx_i, \dd| \btheta),
\end{align}
where \small{$w_i = e^{-\sum_{j=1}^{i-1}\sigma(\xx_j| \btheta)(\tau_{j+1}-\tau_{j})} (1-e^{-\sigma(\xx_i| \btheta)(\tau_{i+1}-\tau_{i})})$}. Intuitively, Equation~\eqref{eq:volume_render} computes the weighted sum of the colors on all sampled points. The weights $w_i$ are computed based on the opacity and the distance to the camera center. 

\mypara{Continual NeRF.} Throughout this paper, we refer to the continual learning problem for NeRFs as \emph{continual NeRF}. At each time step $t$ of continual NeRF: 
\begin{enumerate}
    \item A set of training images along with their camera parameters (intrinsics and extrinsics) $\SS_t$ are generated.
    \item The current model $\btheta_t$ and the replay buffer $\MM_t$ (for storing historical data) are updated by: 
    \begin{align}
        \{\MM_t, \btheta_t\} \leftarrow \text{update}(\SS_t, \btheta_{t-1}, \MM_{t-1})
    \end{align}
    \item $\btheta_t$ is deployed for rendering novel views until $t+1$.
\end{enumerate}
This process simulates the practical scenario where the model $\btheta_t$ is deployed continually. Once in a while, a set of new images arrives, potentially containing new views of the scene and changes in appearance or geometry. The goal is to update $\btheta_t$; ideally storage (to maintain historical data in $\MM_t$) and memory (to deploy $\btheta_t$) requirements are small.

As shown in \cref{fig:overview}, CLNeRF addresses three major problems of continual NeRF: (1) effectively updating $\btheta_t$ using minimal storage, (2) updating $\MM_t$ during optional experience replay, and (3) handling various scene changes with a single compact model. We provide further details on each of these components below.

\subsection{Model Update}\label{sec:model_update}

CLNeRF applies replay-based methods~\cite{er, generative_replay} to prevent catastrophic forgetting. To enable applications with extreme storage limits, CLNeRF combines generative replay~\cite{generative_replay} with advanced NeRF architectures so that it is effective even when no historical image can be stored.

\cref{fig:overview}(a) depicts the model update process of CLNeRF at each time step $t$. The camera parameters of all historical images are stored in the replay buffer $\MM_{t-1}$ for generative replay. A small number of images $\II_{\text{ER}}$ are optionally maintained when the storage is sufficient for experience replay~\cite{er}. 
At each training iteration of $\btheta_t$, CLNeRF generates a batch of camera rays $\xX = \xX_\text{ER} \bigcup \xX_\text{GR} \bigcup \xX_t$ uniformly from $\PP_t \bigcup \PP_\text{ER} \bigcup \PP_\text{GR}$, where $\PP_t$, $\PP_\text{GR}$ and $\PP_\text{ER}$ are respectively the camera parameters of new data $\SS_t$, generative replay data and experience replay data. The training objective is:
\begin{align}
    \underset{\btheta_t}{\text{minimize}} \frac{\sum_{\XX \in \xX}\mathcal{L}_{\text{NeRF}}(C(\XX), \hat{C}(\XX| \btheta_t))}{|\xX|}, 
\end{align}
where $\mathcal{L}_{\text{NeRF}}$ is the loss for standard NeRF training, $C(\cdot)$ is the supervision signal from new data or replay, and $\hat{C}(\cdot|\btheta_t)$ is the color rendered by $\btheta_t$. For the rays $\XX \in \xX_\text{GR}$ sampled from $\PP_\text{GR}$, we perform generative replay, \ie, we set the supervision signal $C(\XX)$ as the image colors $\hat{C}(\XX|\btheta_{t-1})$ generated by $\btheta_{t-1}$. For the other rays, $C(\XX)$ is the ground-truth image color. After the model update, we replace the previously deployed model $\btheta_{t-1}$ with $\btheta_t$ and update the replay buffer $\MM_t$ (see \cref{sec:replay_buffer} for more details). Only $\btheta_t$ and $\MM_t$ are maintained until the next set of data $\SS_{t+1}$ arrives.

Although all camera parameters are stored in $\MM_{t-1}$, they only consume a small amount of storage, at most $N_{t-1}(d_{pose}+d_{int})$, where $N_{t-1}$ is the number of historical images, and $d_{pose}$ and $d_{int}$ are the dimensions of camera poses and intrinsic parameters respectively; $d_{pose}=6$ and $d_{int}\leq 5$ for common camera models~\cite{hartley2003multiple}. $d_{int}$ is shared if multiple images are captured by the same camera. As a concrete example, storing the parameters for 1000 samples each captured with a different camera 
requires roughly $45\text{KB}$ of storage in our experiment, much less than storing a single high resolution image. This guarantees the effectiveness of CLNeRF (see \cref{sec:exp_main}) even for applications with extreme storage limits. 


We also emphasize the importance of random sampling. CLNeRF assigns \emph{uniform} sampling weights between all views revealed so far. Some image-classification-based continual learning methods~\cite{er} and the cocurrent work for NeRFs~\cite{meilnerf} propose biased sampling strategies, where a fixed and large percentage ($\frac{1}{2}$ to $\frac{2}{3}$) of the rays are sampled from new data ($\PP_t$). This strategy not only introduces new hyperparameters (\eg, loss weights of old data or the proportion of rays from new data~\cite{meilnerf}), but also performs worse than uniform random sampling, as shown in \cref{sec:exp}.

\subsection{Replay Buffer Update}\label{sec:replay_buffer}

In the extreme case where no image can be stored for experience replay, we only store the camera parameters of historical data in $\MM_t$ to make CLNeRF flexible and effective for practical systems with various storage sizes. When the storage is sufficient to maintain a subset of historical images for experience replay, 
we use a \emph{reservoir buffer}~\cite{er}. Specifically, current data is added to $\MM_t$ as long as the storage limit is not reached. Otherwise, as shown in \cref{fig:overview} (b), given $\MM_{t-1}$ capable of storing $K$ images, we first generate for each image $\II_i \in \SS_t$ a random integer $j_i \in \{1, 2, ..., N_i\}$, where $N_i$ represents the order of $\II_i$ in the continual learning data sequence. If $j_i > K$, we do not add $\II_i$ into $\MM_t$. Otherwise, we replace the $j_i$'th image in $\MM_{t-1}$ with $\II_i$. Note that $\MM_t$ stores all camera parameters regardless of if the corresponding image is stored or not.

We also experiment with the \emph{prioritized replay buffer}~\cite{prioritized_replay}, where $\MM_t$ keeps images with the lowest rendering quality. Specifically, after updating $\btheta_t$, we iterate over all images in $\MM_{t-1}$ and $\SS_t$ and keep the $K$ images with the lowest rendering PSNR~\cite{psnr_and_ssim} from $\btheta_t$. Though widely used in reinforcement learning, prioritized replay does not perform better than reservoir sampling in continual NeRF (see \cref{sec:exp_repBuf}). A reservoir buffer is also simpler to implement and more efficient (no need to compare the rendering quality) to update; hence CLNeRF applies it by default.

\subsection{Architecture}\label{sec:arch}

CLNeRF by default uses the Instant Neural Graphics Primitives (NGP)~\cite{ngp} architecture. 
This not only enables efficient model updates during continual NeRF, but also ensures the low overhead and effectiveness of generative replay. 
As shown in \cref{sec:exp_arch}, using NGP as the backbone for CLNeRF results in better performance and efficiency compared to vanilla NeRF~\cite{nerf}. 

A compact continual NeRF system should use a single model to incorporate scene changes, so that the model size does not increase significantly over time. We achieve this by adding trainable appearance and geometry embeddings to the base architecture (\cref{fig:overview} (c)). Given a spatial location $\xx$ and a viewing direction $\dd$, we first encode $\xx$ into a feature vector $\ff$ (using the grid-based hash encoder for NGP, and an MLP for vanilla NeRF). Then, we generate the color and opacity respectively by $\cc = D_c(\ff, \dd, \ee_a)$ and $\sigma = D_{\sigma} (\ff, \ee_g)$, where $D_c$ and $D_\sigma$ are the color and opacity decoders (MLP for both NGP and vanilla NeRF); $\ee_a$ is the trainable appearance embedding and $\ee_g$ is the geometry embedding. Given a sequence of scans of the same scene, with appearance and geometry changes between different scans, we add one appearance embedding and one geometry embedding for each scan, i.e., for each time step $t$ of continual NeRF. We set the dimension of appearance and geometry embeddings to 48 and 16 respectively, which ensures minimal model size increase during continual NeRF and is sufficient to encode complex real-world scene changes as shown in \cref{sec:exp}.

\begin{figure}[t]
    \begin{minipage}{.49\columnwidth}
    \centering
    \includegraphics[width=.8\textwidth]{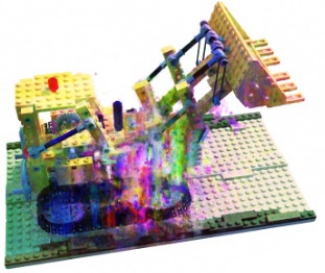}
    \subcaption{Test view rendered by NGP with transient MLP.}
    \end{minipage}
    \begin{minipage}{.49\columnwidth}
    \centering
    \includegraphics[width=.8\textwidth]{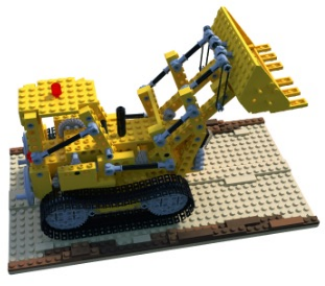}
    \subcaption{Test view rendered by NGP with masked transient objects.}
    \end{minipage}
        \begin{minipage}{.49\columnwidth}
    \centering
    \includegraphics[width=.8\textwidth]{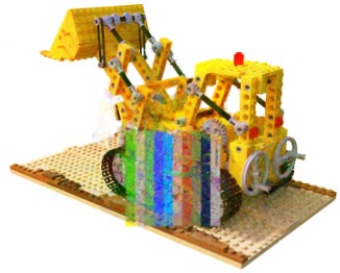}
    \subcaption{Training view rendered by NGP with transient MLP.}
    \end{minipage}
        \begin{minipage}{.49\columnwidth}
    \centering
    \includegraphics[width=.8\textwidth]{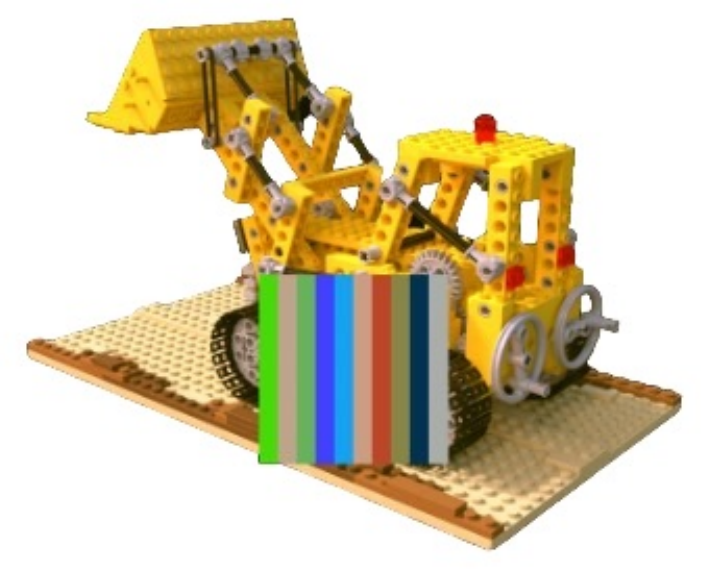}
    \subcaption{GT training view.}
    \end{minipage}
    \vspace{2pt}
    \caption{\textbf{Result of NGP with transient MLPs.} Similar to~\cite{nerfw}, we add artificial transient objects and lightning changes to the \emph{Lego} scene. NGP with transient MLPs overfits to the training views and fails to filter transient objects automatically.}
    \label{fig:Transient}
    \vspace{-2pt}
\end{figure}

CLNeRF uses segmentation masks (from~\cite{deeplab_link}) to filter transient objects. As shown in Fig.~\ref{fig:Transient}, we also explored using the transient MLP and robust training objectives~\cite{nerfw} but found empirically that NGP is not compatible with this strategy -- the non-transient network overfits to the transient objects and fails to filter them automatically. Note that we only remove transient/dynamic objects within \emph{a single scan}, e.g., moving pedestrians. Scene changes between different scans, e.g., newly constructed buildings, are handled by geometry embeddings.

\section{WAT: A Continual NeRF Benchmark}\label{sec:CLRender}





\begin{figure}[t]
    \centering
    \includegraphics[width=1.\columnwidth]{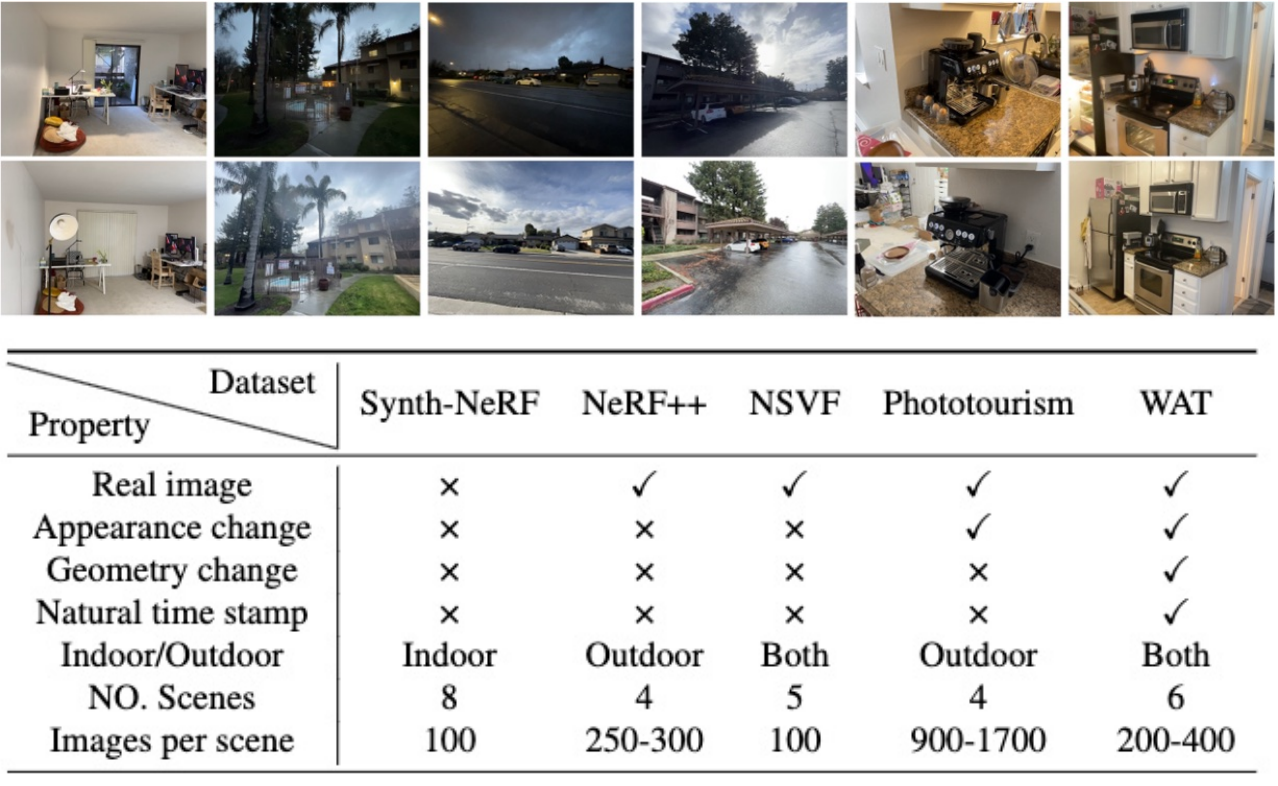}
    \caption{\textbf{Images and properties of the proposed WAT dataset.} WAT contains both indoor and outdoor scenes (samples on top), which change in both appearance and geometry over different scans. Each column of the images shows the same scene scanned at two different times. The scans are naturally ordered according to real-world time during continual NeRF, simulating realistic applications. Compared to WAT, standard benchmarks (see \cref{sec:exp} for details) either lack appearance or geometry change, or natural order of these changes, making continual NeRF less challenging.}
    \label{fig:WAT}
    \vspace{-1em}
\end{figure}



Most continual learning methods in the literature are evaluated on datasets synthesized from standard image classification benchmarks~\cite{cl_survey}. 
 Although a similar strategy can be used on standard NeRF benchmarks, it is not practical as it only considers static scenes with a gradually expanding rendering range. However, this does not model the real-world distribution shifts introduced by the change of time~\cite{cloc, clear_benchmark}, such as the change of scene appearance (\eg, lighting and weather) and geometry (\eg, new decoration of a room). To solve this problem, we propose \emph{World Across Time (WAT)}, a new dataset and benchmark for practical continual NeRF.


As shown in \cref{fig:WAT}, WAT consists of images captured from 6 different scenes (both indoor and outdoor). For each scene, we capture 5-10 videos at different real-world time to generate natural appearance or geometry changes across videos. We extract a subset of the video frames (200-400 images for each scene), and use colmap~\cite{colmap} to compute the camera parameters. For each scene, we hold out $\frac{1}{8}$ of the images for testing and use the remaining images for training. We order the images naturally based on the time that the corresponding videos were captured. At each time step $t$ of continual NeRF, all images belonging to a new video are revealed to the model. Compared to standard NeRF datasets, WAT has diverse scene types, scene changes, and a realistic data order based on real-world time. As shown in \cref{sec:exp_main}, the natural time-based order makes WAT much more challenging than randomly dividing standard in-the-wild datasets into subsets (\eg, as in the case of phototourism, which has similar appearance and pose distributions between different subsets). 
WAT enables us to study the importance of the model design for changing scenes. As shown in \cref{sec:exp_main}, methods designed only for static scenes perform poorly on WAT.

\section{Experiments}\label{sec:exp}

In the experiments, we first compare CLNeRF against other continual learning approaches 
(\cref{sec:exp_main}). Then, we analyse different continual NeRF components in detail (\cref{sec:exp_analysis}). Although NGP is used by default in CLNeRF, we also experiment with vanilla NeRF to demonstrate the effect of architectures.


\mypara{Implementation Details.} Our implementation of the vanilla NeRF and NGP backbones is based on NeRFAcc~\cite{nerfacc} and NGP-PL~\cite{ngp_link} respectively. Following the base implementations, we allow 50K training iterations for vanilla NeRF and 20K for NGP whenever we train or update the model. We find that NGP produces ``NaN" losses and exploding gradients when trained for too long 
, making it hard to initialize $\btheta_t$ with $\btheta_{t-1}$. Hence, we randomly initialize $\btheta_t$ and train the model from scratch at each time step, as done in GDumb~\cite{gdumb}. Empirical results (\cref{sec:exp_arch}) on vanilla NeRF show that initializing $\btheta_t$ with $\btheta_{t-1}$ can help continual NeRF, and we leave further investigation of this issue on NGP for future work.
 Unless otherwise stated, all hyperparameters strictly follow the base code. See Appendix~\ref{appdx:implementation} for further implementation details of different continual learning methods. Training one model with either NGP or vanilla NeRF backbone takes 5-20 minutes or 1-2 hours on a single RTX6000 GPU respectively. 


\mypara{Datasets.} Besides WAT, we also evaluate methods on datasets derived from standard NeRF benchmarks. Specifically, we uniformly divide the training data of standard benchmarks into 10-20 subsets and reveal them sequentially during continual NeRF. For synthetic data, we use the dataset proposed in~\cite{nerf} (referred to as \emph{Synth-NeRF}), resulting in 8 scenes, each with 10 time steps and 20 training images (with consecutive image IDs) per time step. For real-world data, we use two Tanks and Temples~\cite{TaT} subsets proposed in~\cite{nsvf} (with background filter, referred to as \emph{NSVF}) and~\cite{nerf++} (without background filter, referred to as \emph{NeRF++}). Both datasets are divided into multiple sub-trajectories with consecutive video frames. NeRF++ has 4 scenes, 10 time steps and 60-80 images per time step. For NSVF, we mimic the construction process of the concurrent work~\cite{meilnerf}, and divide the first 100 training images of each scene into 20 subsets, resulting in 5 scenes, each with 20 time steps and 5 images per time step. Finally, we use 4 Phototourism scenes (Brandenburg Gate, Sacre Coeur, Trevi Fountain and Taj Mahal) along with the train-test split from~\cite{nerfw}. Due to the lack of time stamps, we randomly divide each scene into 20 time steps and 42-86 images per time step (with consecutive image IDs).

\mypara{Evaluation protocol.} To evaluate a continual NeRF approach, we first train it sequentially over all time steps, then compute the mean PSNR/SSIM~\cite{psnr_and_ssim} on the held-out test images for all time steps. 

\subsection{Main Results}\label{sec:exp_main}

\begin{table*}[t]
\begin{minipage}[t]{0.66\linewidth}
\small{
\centering
\begin{tabular}{c|cccccc}
\toprule
\diagbox{Method}{Dataset} & Synth-NeRF & NeRF++ & NSVF & WAT \\
\midrule
NT (NeRF) & 28.53/0.938 & 14.81/0.462 & 18.58/0.768 & 16.70/0.649 \\ 
EWC (NeRF) & 28.32/0.925 & 15.03/0.442 & 18.12/0.778 & 16.29/0.672 \\ 
ER (NeRF) & 30.29/0.948 & 17.35/0.542 & 26.51/0.902 & 21.36/0.709 \\ 
MEIL-NeRF (NGP) & 30.69/0.953 & 19.40/0.595 & 27.19/0.909 & 24.05/0.730 \\ 
CLNeRF-noER (NGP) & 31.96/0.956 & 20.31/0.632 & 29.32/\textbf{0.924} & 25.44/0.762 \\ 
CLNeRF (NGP) & \textbf{32.16/0.957} & \textbf{20.33/0.634} & \textbf{29.48}/0.923 & \textbf{25.45/0.764} \\  
\midrule
UB (NGP) & 32.94/0.959 & 20.34/0.648 & 30.28/0.931 & 25.85/0.767 \\ 
\bottomrule
\end{tabular}
}
\end{minipage}
\begin{minipage}[t]{0.33\linewidth}
\centering
\small{
\begin{tabular}{c|c}
\toprule
  \diagbox{Method}{Dataset} & Phototourism \\
\midrule
NT (NGP) & 19.28/0.692 \\ 
ER (NGP) & 20.03/0.713 \\ 
MEIL-NeRF (NGP) & 22.35/0.746 \\ 
CLNerf-noER (NGP) & 22.67/0.751 \\ 
CLNerf (NGP) & \textbf{22.88/0.752} \\  
\midrule
UB (NeRFW) & 22.78/0.823 \\ 
UB (NGP) & 23.05/0.763 \\ 
\bottomrule
\end{tabular}
}
\end{minipage}
\vspace{3pt}
\caption{\textbf{Main results}. The results are in the form of PSNR/SSIM~\cite{psnr_and_ssim}, with the best performing method in bold. We label each method with the best performing architecture, \ie, (vanilla) NeRF or NGP. 
CLNeRF performs the best among all continual NeRF approaches and across all datasets, even without storing any historical images (CLNeRF-noER). The performance gap between CLNeRF and the upper-bound model UB remains low for all datasets. We equip all competitors on WAT with trainable embeddings proposed in \cref{sec:arch} for fairer comparison to CLNeRF. Without using the embeddings, the performance gap between these methods and CLNeRF increases significantly (NT: 16.17/0.625, EWC: 15.81/0.650, ER: 17.99/0.666, MEIL-NeRF: 20.92/0.720). On Phototourism, the performance difference across methods is much smaller due to the random division of time steps (making pose and appearance distributions similar over time). CLNerf still performs close to the upper-bound model. UB with NerfW performs worse than UB with NGP in terms of PSNR but better in terms of SSIM.}
\label{tab:main}
\end{table*}

To evaluate CLNeRF, we compare it against: (1) \emph{Naive Training (NT)}, where we train a model sequentially on new data without continual learning. NT represents the lower-bound performance that we can achieve on continual NeRF. (2) \emph{Elastic Weight Consolidation (EWC)}~\cite{ewc}, a widely-used regularization-based continual learning method. (3) \emph{Experience Replay (ER)}~\cite{er}, one of the most effective continual learning methods. (4) \emph{MEIL-NeRF}~\cite{meilnerf}, a concurrent work that also uses generative replay. For fair comparison, we use the ground-truth camera parameters to generate replay camera rays for MEIL-NeRF as done in CLNeRF, rather than using a small MLP to learn the rays of interests. This strategy makes the implementation simpler and performs better, as also demonstrated in the original paper. (5) The \emph{upper bound model (UB)} trained on all data at once, representing the upper-bound performance of continual NeRF. For all methods that involve experience replay (ER and CLNeRF), we allow 10 images to be stored in the replay buffer to simulate the case of highly limited storage (see \cref{sec:exp_repBuf} for the effect of the replay buffer size). For fair comparison, we choose the best-performing architecture for each method. See \cref{sec:exp_analysis} for the effect of architectures, and Appendix~\ref{appdx:ngp_baseline} for NGP-only results.

As shown in \cref{tab:main}, CLNeRF, even without storing historical data for experience replay (CLNeRF-noER), performs much better than other continual NeRF approaches across all datasets (see Appendix~\ref{appdx:more_qualitative} for the results of individual scenes). With only 10 images (2.5\%-10\% of the complete dataset size) stored in the replay buffer, CLNeRF achieves comparable performance as UB, which requires storing all historical data. Although MEIL-NeRF also applies generative replay, the biased sampling strategy (towards new data) and the complex loss designed for vanilla NeRF are not suitable for advanced architectures like NGP. As a result, there is a significant performance gap compared to UB which is consistent with the results in the original paper. As shown later in \cref{sec:exp_arch}, CLNeRF also performs better than MEIL-NeRF with a vanilla NeRF backbone. Interestingly, methods without generative replay (NT, EWC, ER) work better with vanilla NeRF. We analyze this phenomenon in detail in \cref{sec:exp_arch}. For a fairer comparison on WAT, we use the trainable embeddings of CLNeRF also for the other methods in Tab.~\ref{tab:main}. The results without embeddings are reported in the caption; the gap to CLNeRF increases significantly. 


\begin{figure*}[t]
	\center
	\includegraphics[width=1.\textwidth]{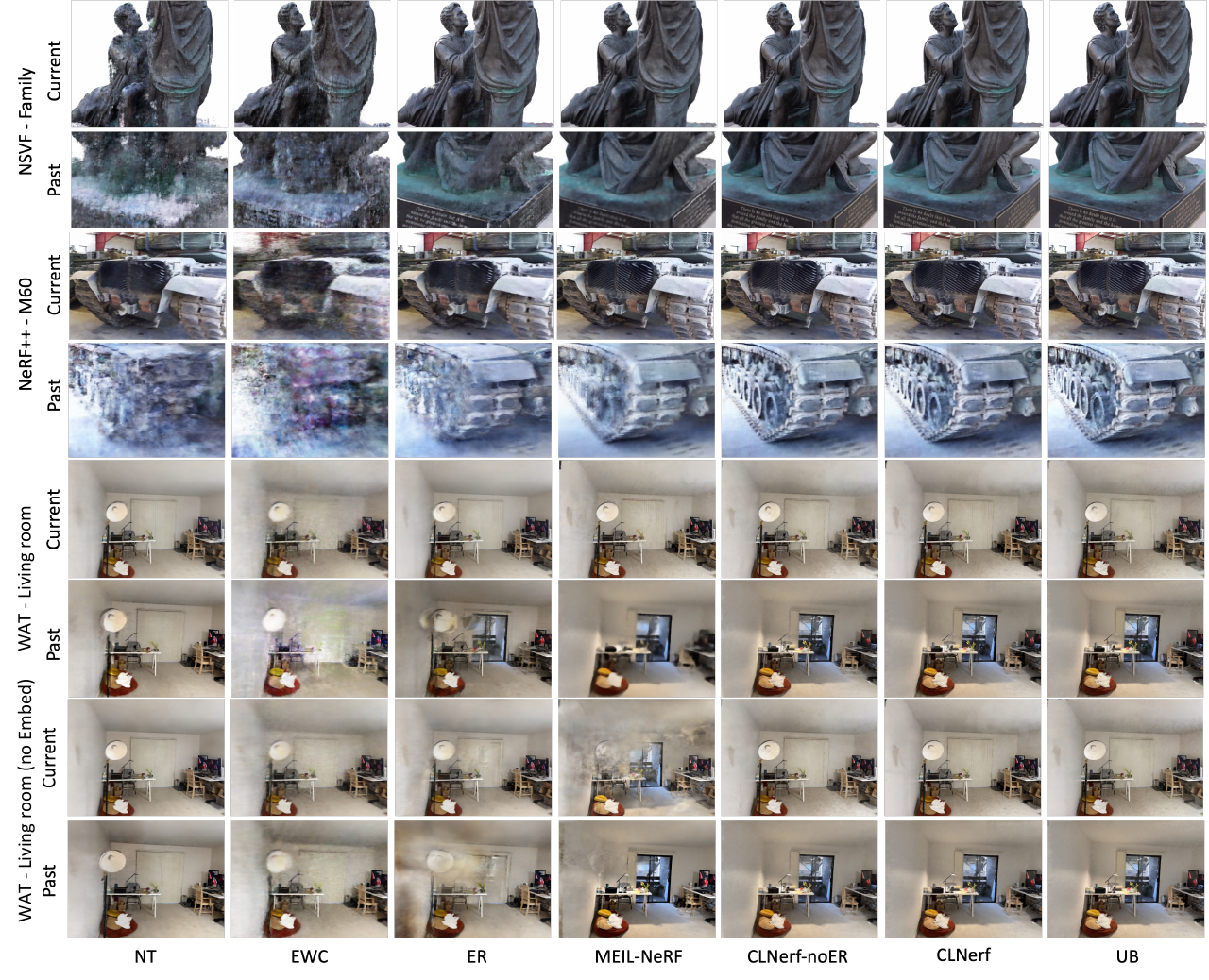}
 \vspace{-1.7em}
	\caption{\textbf{Qualitative results.} Each two rows show the (zoom-in) test views of the current and past scans rendered by different methods. CLNeRF has a similar rendering quality as UB, even without storing any historical images. NT overfits to the new data, resulting in erroneous renderings for early scans. The regularization from EWC not only hinders the model from adapting to new data but also fails to recover the old scene appearance/geometry. Blur and artifacts appear on images rendered by ER and MEIL-NeRF, especially in early scans, due to the lack of enough replay data (ER), the biased sampling and loss function design (MEIL-NeRF). Without using the trainable embeddings proposed in CLNeRF (WAT - Living room (noEmbed)), other continual NeRF approaches perform much worse on WAT.}\label{fig:qualitative}
  \vspace{-1em}
\end{figure*}

We also apply CLNeRF to in-the-wild images from Phototourism. Since the NeRFAcc-based vanilla NeRF implementation does not perform well on Phototourism, and NeRFW~\cite{nerfw} is slow ($>$ 1 week per time step with 8 GPUs) in continual NeRF, we instead report the performance using NGP for NT and ER, and the UB version of NeRFW based on the implementation of~\cite{nerfw_pl}. EWC cannot be applied to NGP since we need to perform re-initialization at each time step. We assign 1 appearance embedding to each image (rather than each time step) to handle per-image appearance change.  As shown in \cref{tab:main} the NGP-based upper bound model performs better than NeRFW in terms of PSNR and worse in terms of SSIM. CLNeRF performs close to the upper bound model. Due to the lack of time stamps, we do not have a natural data order for Phototourism. This simplifies the problem since images from different (artificially created) time steps have similar pose and appearance distributions. As a result, the performance gap between different methods is much smaller compared to other datasets, even with a much larger number of images (800-1700 in Phototourism versus 100-400 in other datasets). 

\cref{fig:qualitative} shows qualitative results (see Appendix~\ref{appdx:more_qualitative} for more). Each two rows show the novel views rendered for the current and past time steps. CLNeRF provides similar rendering quality as UB, with a much lower storage consumption. Without using continual learning, NT overfits to the current data, resulting in wrong geometry (the redundant lamp and the wrongly closed curtain in \emph{Living room - past}), lightning and severe artifacts for past time steps. Due to the lack of historical data, EWC not only fails to recover past scenes, but also hinders the model from learning on new data. ER stores a small amount of historical data to prevent forgetting. However, the limited replay buffer size makes it hard to recover details in past time steps (see \cref{sec:exp_repBuf} for the effect of replay buffer sizes). The biased sampling strategy and complex loss design make MEIL-NeRF not only more complex (\eg, extra hyperparameters), but also underfit on historical data. As a result, it loses detail from past time steps, even when equipped with the same trainable embeddings of CLNeRF. 

These results show the importance of WAT on benchmarking continual NeRF under practical scene changes. They also show the superiority of CLNeRF over other baselines to robustly handle appearance and geometry changes.



\subsection{Analysis}\label{sec:exp_analysis}
\subsubsection{Ablation}

\begin{table}[t]
\centering
\small{
\begin{tabular}{c|cc}
\toprule
  \diagbox{Method}{Dataset} & Synth-NeRF  & WAT \\
\midrule
CLNeRF & \textbf{32.16/0.957} & \textbf{25.45/0.764} \\ 
No ER & 31.96/0.957 & 25.44/0.762 \\ 
No GenRep & 27.35/0.919 & 18.52/0.634 \\ 
No NGP & 29.23/0.940 & 23.50/0.728 \\ 
No Embed & N.A. & 21.09/0.725 \\
\bottomrule
\end{tabular}
}
\vspace{3pt}
\caption{\textbf{Ablation study}. The performance of CLNeRF drops slightly without ER, but significantly without generative replay (No GenRep) or the use of the NGP architecture. Without using the trainable embeddings (No Embed), CLNeRF performs much worse under the appearance and geometry changes of WAT. Synth-NeRF has only static scenes, hence no trainable embeddings are required. 
}\label{tab:ablation}
\end{table}

This section analyzes the effectiveness of individual CLNeRF components. Specifically, we remove each component and report the performance drop. As shown in Tab.~\ref{tab:ablation}, the performance drops only slightly without experience replay (No ER). However, without generative replay (No GenRep), the performance dropps significantly. Note that NoGenRep is ER with NGP instead of vanilla NeRF. Hence, generative replay is more important in CLNeRF than experience replay to prevent forgetting. Without using NGP, \ie, when applied to vanilla NeRF, CLNeRF also performs much worse, and sometimes (\eg, on Synth-NeRF) worse than ER (NeRF) in \cref{tab:main}. This result shows the importance of advanced architectures for guaranteeing the effectiveness of generative replay. Without the trainable embeddings (No Embed), CLNeRF cannot adapt well to changing appearance and geometry of WAT. 

\subsubsection{Effect of Replay Buffer}\label{sec:exp_repBuf}


\begin{figure}[t]
	\center
	\includegraphics[width=1.\columnwidth]{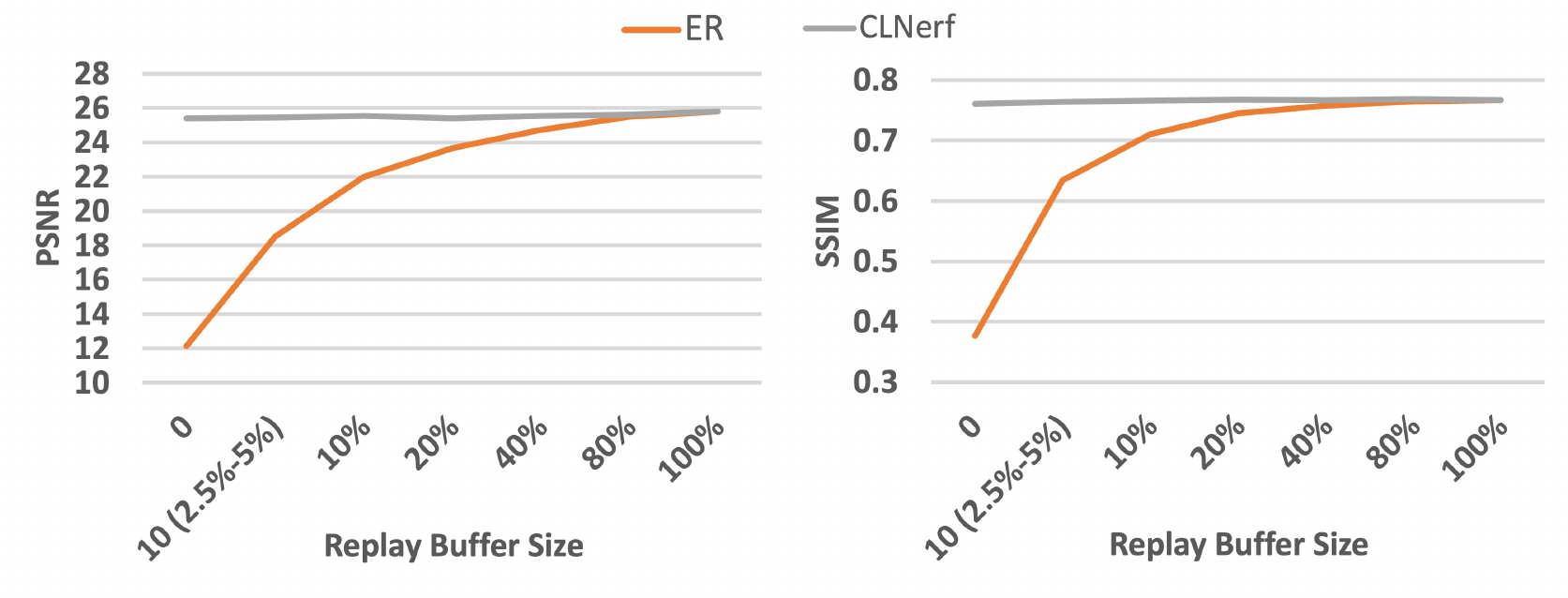}
	\caption{\textbf{Effect of replay buffer sizes (on WAT).} The performance of CLNeRF remains high across different replay buffer sizes. ER on NGP requires a replay buffer size of more than $80\%$ of the dataset size to perform on-par with CLNeRF. Note that $80\%$ corresponds to almost all historical data before the current time step, since all data from the current time step is always available. ``$10 (2.5\%-5\%)$" means we allow 10 images to be stored in the replay buffer, which is roughly $2.5\%-5\%$ of all images of a scene. ``$10\%$" means the replay buffer can store ``10\%" of all images from the same scene.}\label{fig:repBufSize}
\end{figure}

\mypara{Replay Buffer Size.} To mimic the case of highly limited storage, we only allow 10 historical images to be stored for experience replay in the main experiment. Here, we investigate the effect of replay buffer size. Specifically, we vary the replay buffer size of ER (with NGP) and CLNeRF in the pattern of $\{0, 10, 10\%, ... , 100\%\}$ and report the performance change. 0 and 10 (roughly $2.5\%-5\%$ on WAT) are the number of stored images. The percentages are with respect to all images across time steps. As shown in \cref{fig:repBufSize}, CLNeRF does not require any samples stored in the replay buffer to perform well but it also does not hurt performance. ER requires a large replay buffer size ($80\%$) to perform on-par with CLNeRF. This interesting result shows that widely-used CL methods designed for image classification can be sub-optimal for other problems.






\begin{table}[t]
\centering
\small{
\begin{tabular}{c|cc}
\toprule
  \diagbox{Method}{Dataset} & Synth-NeRF  & WAT \\
\midrule
Reservoir & 32.16/0.957 & 25.45/0.764 \\
Prioritized & 32.16/0.957 & 25.40/0.767 \\ \bottomrule
\end{tabular}
}
\vspace{3pt}
\caption{\textbf{Effect of replay buffer update methods}. Prioritized sampling and reservoir sampling perform similarly. Due to the simplicity and efficiency, we use a reservoir buffer for CLNeRF.}
\label{tab:repBufType}
\end{table}

\mypara{Replay Buffer Update Strategy.} CLNeRF applies reservoir sampling to update the replay buffer at each time step. Here, we analyze the effect of different replay buffer update strategies. Specifically, we compare the performance of CLNeRF using reservoir sampling~\cite{er} and prioritized replay~\cite{prioritized_replay}. As shown in Tab.~\ref{tab:repBufType}, changing the reservoir buffer to a prioritized replay buffer does not improve CLNeRF. Hence, a uniform coverage of the whole scene (changes) is sufficient for effective experience replay.

\subsubsection{Effect of Architecture}\label{sec:exp_arch}

\begin{table}[t]
\centering
\small{
\begin{tabular}{c|cc}
\toprule
  \diagbox{Method}{Dataset} & SynthNeRF & WAT \\
\midrule
MEIL-NeRF (NeRF) & 27.99/0.931 & 23.05/0.721 \\ 
CLNeRF-noER (NeRF) & 28.56/0.936 & 23.40/0.727 \\ 
UB (NeRF) & 31.52/0.948 & 24.01/0.740 \\ \bottomrule
\end{tabular}
}
\vspace{3pt}
\caption{\textbf{CLNeRF vs. MEIL with vanilla NeRF}. CLNeRF also outperforms MEIL-NeRF with vanilla NeRF backbone, despite equipping MEIL-NeRF with the proposed trainable embeddings.}\label{tab:CL_MEIL_NeRF}
\end{table}

To show the effectiveness of CLNeRF across architectures, we compare it against UB and MEIL-NeRF using vanilla NeRF as a backbone. We do not use experience replay for either CLNeRF or MEIL-NeRF, and we equip all methods with the trainable embeddings proposed in \cref{sec:arch}. As shown in \cref{tab:CL_MEIL_NeRF}, CLNeRF still performs slightly better than MEIL-NeRF, even though MEIL-NeRF was specifically designed based on vanilla NeRF. The performance gap between CLNeRF/MEIL-NeRF and UB on SynthNeRF is larger with vanilla NeRF than with NGP, highlighting the importance of advanced architectures for generative replay.

\begin{table}[t]
\centering
\small{
\begin{tabular}{c|ccc}
\toprule
 Method & NeRF & NeRF-Reinit & NGP \\
\midrule
NT & 28.53/0.938 & 25.01/0.900  & 21.66/0.858 \\ 
ER & 30.29/0.948 & 28.09/0.928 & 27.35/0.919 \\ 
CLNeRF-noER & 28.56/0.936 & 28.20/0.933 & 31.96/0.956 \\
\midrule
UB & 31.52/0.948 & NA & 32.94/0.959 \\ 
\bottomrule
\end{tabular}
}
\vspace{3pt}
\caption{\textbf{Effect of architectures to different methods (on Synth-NeRF)}. The performance gain of NT and ER using vanilla NeRF comes largely from the capability to initialize the current model $\btheta_t$ with the previous model $\btheta_{t-1}$. With re-initialization, vanilla NeRF still performs better than NGP on ER and NT. We conjecture that NGP overfits more to training data in the case of sparse views. Generative replay in CLNeRF allows NGP to overcome this issue, and perform better than vanilla NeRF.}\label{tab:effect_arch}
\end{table}

As shown in Tab.~\ref{tab:main}, both ER and NT benefit more from vanilla NeRF. To reveal the underlying reason, we compare NT, ER, CLNeRF under 3 different training strategies: (1) Trained using vanilla NeRF without re-initialization (NeRF). (2) Trained using vanilla NeRF with re-initialization at each time step $t$ (NeRF-Reinit). (3) Trained using NGP (NGP). As shown in \cref{tab:effect_arch}, a large portion of the performance gap lies in the inheritance of model parameters from previous time steps, \ie, not performing re-initialization for vanilla NeRF. When both are re-initialized, vanilla NeRF still performs slightly better than NGP for methods without generative replay. We conjecture that this is because NGP overfits more to the training data given sparse views (which is the case for NT and ER), and generalizes poorly on novel views. Performing generative replay allows NGP to overcome the sparse training view issue and exceed the performance of vanilla NeRF.

\section{Conclusion}

This work studies continual learning for NeRFs. We propose a new dataset -- World Across Time (WAT) -- containing natural scenes with appearance and geometry changes over time. We also propose CLNeRF, an effective continual learning system that performs close to the upper bound model trained on all data at once. CLNeRF uses generative replay and performs well even without storing any historical images. While our current experiments only cover scenes with hundreds of images, they are an important step toward deploying practical NeRFs in the real world. There are many interesting future research directions for CLNeRF. For example, solving the NaN loss problem of NGP to make model inheritance more effective during continual learning. Extending CLNeRF to the scale of Block-NeRF~\cite{blocknerf} is also an interesting future work.

{\small
\bibliographystyle{ieee_fullname}
\bibliography{egbib}

\begin{thebibliography}{10}\itemsep=-1pt

\bibitem{nerfw_pl}
Nerfw pytorch lightning implementation.
\newblock "\url{https://github.com/kwea123/nerf_pl/tree/nerfw}".

\bibitem{ngp_link}
{NGP pytorch lightning}.
\newblock \url{https://github.com/kwea123/ngp_pl}.

\bibitem{deeplab_link}
{pretrained deeplabv3 model}.
\newblock
  \url{https://pytorch.org/vision/main/models/generated/torchvision.models.segmentation.deeplabv3_resnet101.html##torchvision.models.segmentation.DeepLabV3_ResNet101_Weights}.

\bibitem{cloc}
Zhipeng Cai, Ozan Sener, and Vladlen Koltun.
\newblock Online continual learning with natural distribution shifts: An
  empirical study with visual data.
\newblock In {\em Proceedings of the IEEE/CVF International Conference on
  Computer Vision}, pages 8281--8290, 2021.

\bibitem{er}
Arslan Chaudhry, Marcus Rohrbach, Mohamed Elhoseiny, Thalaiyasingam Ajanthan,
  Puneet~K Dokania, Philip~HS Torr, and Marc'Aurelio Ranzato.
\newblock On tiny episodic memories in continual learning.
\newblock {\em arXiv preprint arXiv:1902.10486}, 2019.

\bibitem{meilnerf}
Jaeyoung Chung, Kanggeon Lee, Sungyong Baik, and Kyoung~Mu Lee.
\newblock Meil-nerf: Memory-efficient incremental learning of neural radiance
  fields.
\newblock {\em arXiv preprint arXiv:2212.08328}, 2022.

\bibitem{cl_survey}
Matthias De~Lange, Rahaf Aljundi, Marc Masana, Sarah Parisot, Xu Jia,
  Ale{\v{s}} Leonardis, Gregory Slabaugh, and Tinne Tuytelaars.
\newblock A continual learning survey: Defying forgetting in classification
  tasks.
\newblock {\em IEEE transactions on pattern analysis and machine intelligence},
  44(7):3366--3385, 2021.

\bibitem{nerf_survey}
Kyle Gao, Yina Gao, Hongjie He, Denning Lu, Linlin Xu, and Jonathan Li.
\newblock Nerf: Neural radiance field in 3d vision, a comprehensive review.
\newblock {\em arXiv preprint arXiv:2210.00379}, 2022.

\bibitem{gan}
Ian Goodfellow, Jean Pouget-Abadie, Mehdi Mirza, Bing Xu, David Warde-Farley,
  Sherjil Ozair, Aaron Courville, and Yoshua Bengio.
\newblock Generative adversarial networks.
\newblock {\em Communications of the ACM}, 63(11):139--144, 2020.

\bibitem{hartley2003multiple}
Richard Hartley and Andrew Zisserman.
\newblock {\em Multiple view geometry in computer vision}.
\newblock Cambridge university press, 2003.

\bibitem{psnr_and_ssim}
Alain Hore and Djemel Ziou.
\newblock Image quality metrics: Psnr vs. ssim.
\newblock In {\em 2010 20th international conference on pattern recognition},
  pages 2366--2369. IEEE, 2010.

\bibitem{ewc}
James Kirkpatrick, Razvan Pascanu, Neil Rabinowitz, Joel Veness, Guillaume
  Desjardins, Andrei~A Rusu, Kieran Milan, John Quan, Tiago Ramalho, Agnieszka
  Grabska-Barwinska, et~al.
\newblock Overcoming catastrophic forgetting in neural networks.
\newblock {\em Proceedings of the national academy of sciences},
  114(13):3521--3526, 2017.

\bibitem{TaT}
Arno Knapitsch, Jaesik Park, Qian-Yi Zhou, and Vladlen Koltun.
\newblock Tanks and temples: Benchmarking large-scale scene reconstruction.
\newblock {\em ACM Transactions on Graphics (ToG)}, 36(4):1--13, 2017.

\bibitem{nerfacc}
Ruilong Li, Matthew Tancik, and Angjoo Kanazawa.
\newblock Nerfacc: A general nerf acceleration toolbox.
\newblock {\em arXiv preprint arXiv:2210.04847}, 2022.

\bibitem{lwf}
Zhizhong Li and Derek Hoiem.
\newblock Learning without forgetting.
\newblock {\em IEEE transactions on pattern analysis and machine intelligence},
  40(12):2935--2947, 2017.

\bibitem{clear_benchmark}
Zhiqiu Lin, Jia Shi, Deepak Pathak, and Deva Ramanan.
\newblock The clear benchmark: Continual learning on real-world imagery.
\newblock In {\em Thirty-fifth Conference on Neural Information Processing
  Systems Datasets and Benchmarks Track (Round 2)}, 2021.

\bibitem{nsvf}
Lingjie Liu, Jiatao Gu, Kyaw Zaw~Lin, Tat-Seng Chua, and Christian Theobalt.
\newblock Neural sparse voxel fields.
\newblock {\em Advances in Neural Information Processing Systems},
  33:15651--15663, 2020.

\bibitem{gem}
David Lopez-Paz and Marc'Aurelio Ranzato.
\newblock Gradient episodic memory for continual learning.
\newblock {\em Advances in neural information processing systems}, 30, 2017.

\bibitem{packnet}
Arun Mallya and Svetlana Lazebnik.
\newblock Packnet: Adding multiple tasks to a single network by iterative
  pruning.
\newblock In {\em Proceedings of the IEEE conference on Computer Vision and
  Pattern Recognition}, pages 7765--7773, 2018.

\bibitem{nerfw}
Ricardo Martin-Brualla, Noha Radwan, Mehdi~SM Sajjadi, Jonathan~T Barron,
  Alexey Dosovitskiy, and Daniel Duckworth.
\newblock Nerf in the wild: Neural radiance fields for unconstrained photo
  collections.
\newblock In {\em Proceedings of the IEEE/CVF Conference on Computer Vision and
  Pattern Recognition}, pages 7210--7219, 2021.

\bibitem{volume_render_survey}
Nelson Max.
\newblock Optical models for direct volume rendering.
\newblock {\em IEEE Transactions on Visualization and Computer Graphics},
  1(2):99--108, 1995.

\bibitem{catastrophic_forgetting}
Michael McCloskey and Neal~J Cohen.
\newblock Catastrophic interference in connectionist networks: The sequential
  learning problem.
\newblock In {\em Psychology of learning and motivation}, volume~24, pages
  109--165. Elsevier, 1989.

\bibitem{nerf}
Ben Mildenhall, Pratul~P Srinivasan, Matthew Tancik, Jonathan~T Barron, Ravi
  Ramamoorthi, and Ren Ng.
\newblock Nerf: Representing scenes as neural radiance fields for view
  synthesis.
\newblock {\em Communications of the ACM}, 65(1):99--106, 2021.

\bibitem{ngp}
Thomas M{\"u}ller, Alex Evans, Christoph Schied, and Alexander Keller.
\newblock Instant neural graphics primitives with a multiresolution hash
  encoding.
\newblock {\em ACM Transactions on Graphics (ToG)}, 41(4):1--15, 2022.

\bibitem{gdumb}
Ameya Prabhu, Philip~HS Torr, and Puneet~K Dokania.
\newblock Gdumb: A simple approach that questions our progress in continual
  learning.
\newblock In {\em Computer Vision--ECCV 2020: 16th European Conference,
  Glasgow, UK, August 23--28, 2020, Proceedings, Part II 16}, pages 524--540.
  Springer, 2020.

\bibitem{kilonerf}
Christian Reiser, Songyou Peng, Yiyi Liao, and Andreas Geiger.
\newblock Kilonerf: Speeding up neural radiance fields with thousands of tiny
  mlps.
\newblock In {\em Proceedings of the IEEE/CVF International Conference on
  Computer Vision}, pages 14335--14345, 2021.

\bibitem{prioritized_replay}
Tom Schaul, John Quan, Ioannis Antonoglou, and David Silver.
\newblock Prioritized experience replay.
\newblock {\em arXiv preprint arXiv:1511.05952}, 2015.

\bibitem{colmap}
Johannes~Lutz Sch\"{o}nberger and Jan-Michael Frahm.
\newblock Structure-from-motion revisited.
\newblock In {\em Conference on Computer Vision and Pattern Recognition
  (CVPR)}, 2016.

\bibitem{HAT}
Joan Serra, Didac Suris, Marius Miron, and Alexandros Karatzoglou.
\newblock Overcoming catastrophic forgetting with hard attention to the task.
\newblock In {\em International conference on machine learning}, pages
  4548--4557. PMLR, 2018.

\bibitem{generative_replay}
Hanul Shin, Jung~Kwon Lee, Jaehong Kim, and Jiwon Kim.
\newblock Continual learning with deep generative replay.
\newblock {\em Advances in neural information processing systems}, 30, 2017.

\bibitem{sun2022direct}
Cheng Sun, Min Sun, and Hwann-Tzong Chen.
\newblock Direct voxel grid optimization: Super-fast convergence for radiance
  fields reconstruction.
\newblock In {\em Proceedings of the IEEE/CVF Conference on Computer Vision and
  Pattern Recognition}, pages 5459--5469, 2022.

\bibitem{blocknerf}
Matthew Tancik, Vincent Casser, Xinchen Yan, Sabeek Pradhan, Ben Mildenhall,
  Pratul~P Srinivasan, Jonathan~T Barron, and Henrik Kretzschmar.
\newblock Block-nerf: Scalable large scene neural view synthesis.
\newblock In {\em Proceedings of the IEEE/CVF Conference on Computer Vision and
  Pattern Recognition}, pages 8248--8258, 2022.

\bibitem{nerf++}
Kai Zhang, Gernot Riegler, Noah Snavely, and Vladlen Koltun.
\newblock Nerf++: Analyzing and improving neural radiance fields.
\newblock {\em arXiv preprint arXiv:2010.07492}, 2020.

\end{thebibliography}
}

\clearpage

\appendix




\section{Further Implementation Details for Different Methods}\label{appdx:implementation}

\noindent \textbf{EWC.} EWC~\cite{ewc} regularizes the current model $\btheta_t$ to be close to $\btheta_{t-1}$. The training loss of EWC at time step $t$ is
\begin{align}
    \mathcal{L}_{\text{EWC}}(\btheta_t) = \mathcal{L}_{\text{NeRF}}(\btheta_t) + \sum_i \frac{\lambda}{2} F_i (\btheta_{t,i} - \btheta_{t-1,i})^2,
\end{align}
where $\mathcal{L}_{\text{NeRF}}(\cdot)$ is the original NeRF training loss, $F_i$ is the i-th element of the diagnal of the Fisher information matrix, $\btheta_{t,i}$ is the i-th element of $\btheta_t$, and $\lambda$ is the hyper-parameter that controls the regularization strength. As in the original paper, the regularization term from EWC has a much smaller magnitude than $\mathcal{L}_{\text{NeRF}}$. To make regularization effective, we tune $\lambda$ by grid search from $1\mathrm{e}3$ to $1\mathrm{e}{9}$ on the Synth-NeRF dataset, and picking $1\mathrm{e}{5}$ that provides the best performance (making the regularization term roughly 10\% of $\mathcal{L}_{\text{NeRF}}$). 

\noindent\textbf{ER.} Experience replay (ER)~\cite{er} uses the loss on both new and historical data to update the model $\btheta_t$. Due to the ineffectiveness of biased sampling for continual NeRF, we uses random sampling for ER to produce the best performance and uniformly weight the losses of all ray samples.

\noindent\textbf{MEIL-NeRF.} MEIL-NeRF~\cite{meilnerf} forms each mini-batch of training data by sampling $\frac{2}{3}$ of the rays from new data, and the rest from old ones. The training loss of MEIL-NeRF is
\begin{align}\label{eq:MEIL}
\resizebox{.9\hsize}{!}{$\frac{1}{|\XX_c|} \sum_{\XX_c} \mathcal{L}_{\text{NeRF}}(\XX_c, \btheta_t) + \frac{\lambda_{\text{MEIL}}}{|\XX_o|} \sum_{\XX_o} \rho (\hat{C} (\XX_o|\btheta_{t-1}) - \hat{C} (\XX_o|\btheta_t))$},
\end{align}
where $\XX_c$ and $\XX_o$ are respectively the rays from new and old data, $\mathcal{L}_{\textbf{NeRF}}(\cdot)$ is the original loss for NeRF training, $\rho(\cdot)$ is the Charbonnier penalty function, $\hat{C}(\XX|\btheta)$ is the color of ray $\XX$ generated by $\btheta$, and $\lambda_{\text{MEIL}}$ is the hyper-parameter that controls the regularization strength from old data. Following S2 of the original paper, $\lambda_{\text{MEIL}}$ is scheduled as 
\begin{align}
 \lambda_{\text{MEIL}} = \frac{\cos (\pi (1+r)) + 1}{2},   
\end{align}
where $r$ is the training progress rate (from 0 to 1) in each time step. Note that we correct the typo of the original paper and add a ``$+1$'' after $\cos(\cdot)$ to ensure that $\lambda$ grows gradually from 0 to 1 (consistent with Figure S5 of the original paper).

\noindent\textbf{CLNeRF} Depending on the resource limit, the generative replay of CLNeRF can be implemented in both online and offline fashion. For applications where new data are generated only once a while (e.g., city scans are uploaded once a couple of days), the model is mostly in the deploy mode. Hence during the infrequent case of model update, we can assign a temporal storage to store all images generated by $\btheta_{t-1}$, and then use them to update $\btheta_t$. The benefit of this implementation is that we only need to load 1 model into the GPU memory, and the temporal storage can be released once we finish updating $\btheta_t$ (which takes only 5-20 mins for CLNeRF). For applications where no temporal storage can be used, we load $\btheta_{t-1}$ (evaluation mode to same memory) and $\btheta_t$ into the GPU at the same time, and generate the replay supervision signal on-the-fly, which requires an extra forward pass per training iteration. In our test, such implementation can still fit into a single RTX6000 GPU, and increased the training time by roughly $60\%$, which is still fast for NGP. We use the first implementation in our experiments due to its simplicity.

\section{Further Qualitative Results on WAT} \label{appdx:more_qualitative}

\begin{figure*}[t]
	\center
	\includegraphics[width=1.\textwidth]{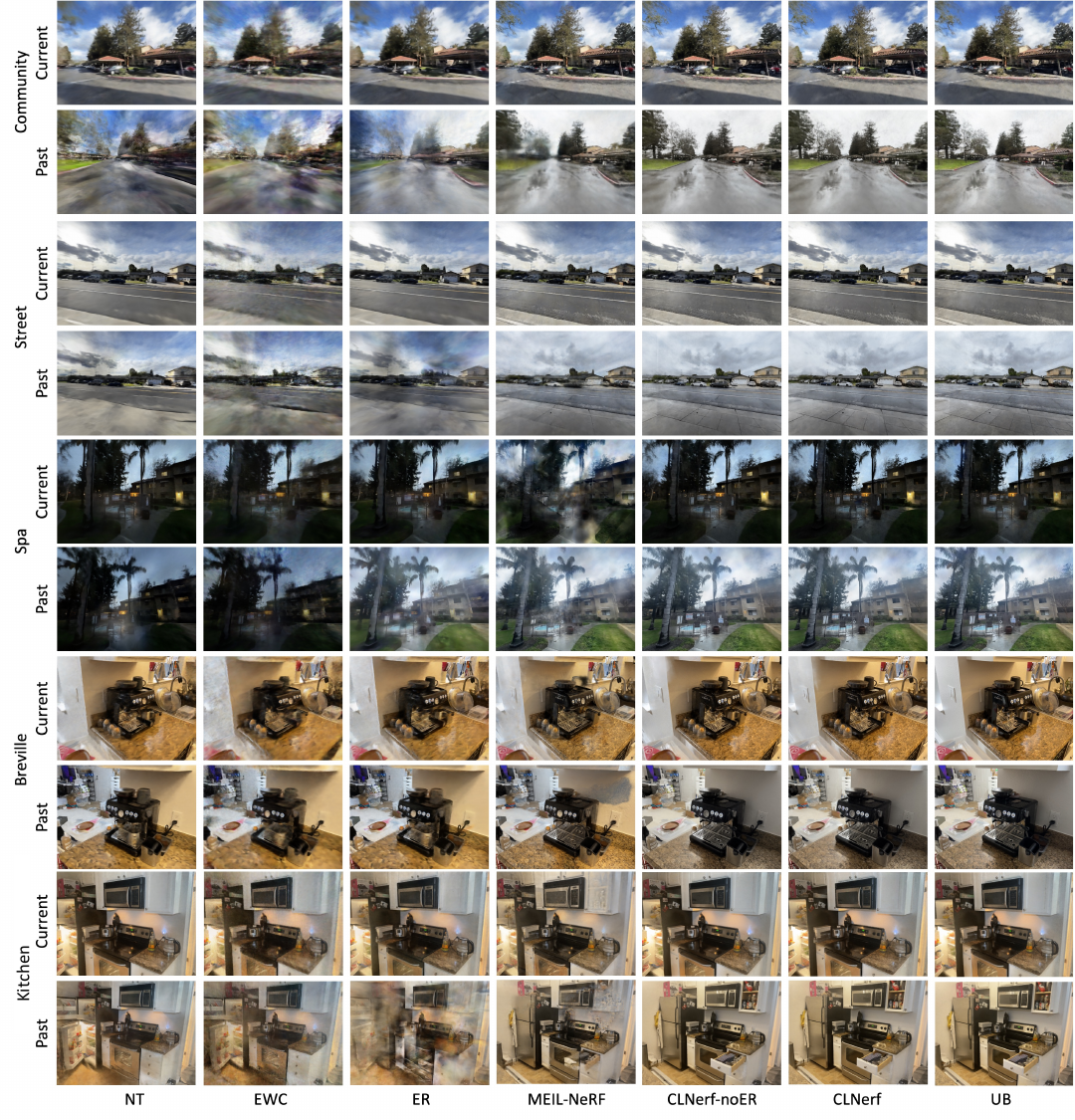}
	\caption{\textbf{Further qualitative results without using trainable embeddings for other continual NeRF baselines (zoom-in recommended).} Methods without trainable embeddings cannot properly recover the appearance and geometry of the scene at past time steps.}\label{fig:more_qualitative_noEmbed}
\end{figure*}

\begin{figure*}[t]
	\center
	\includegraphics[width=1.\textwidth]{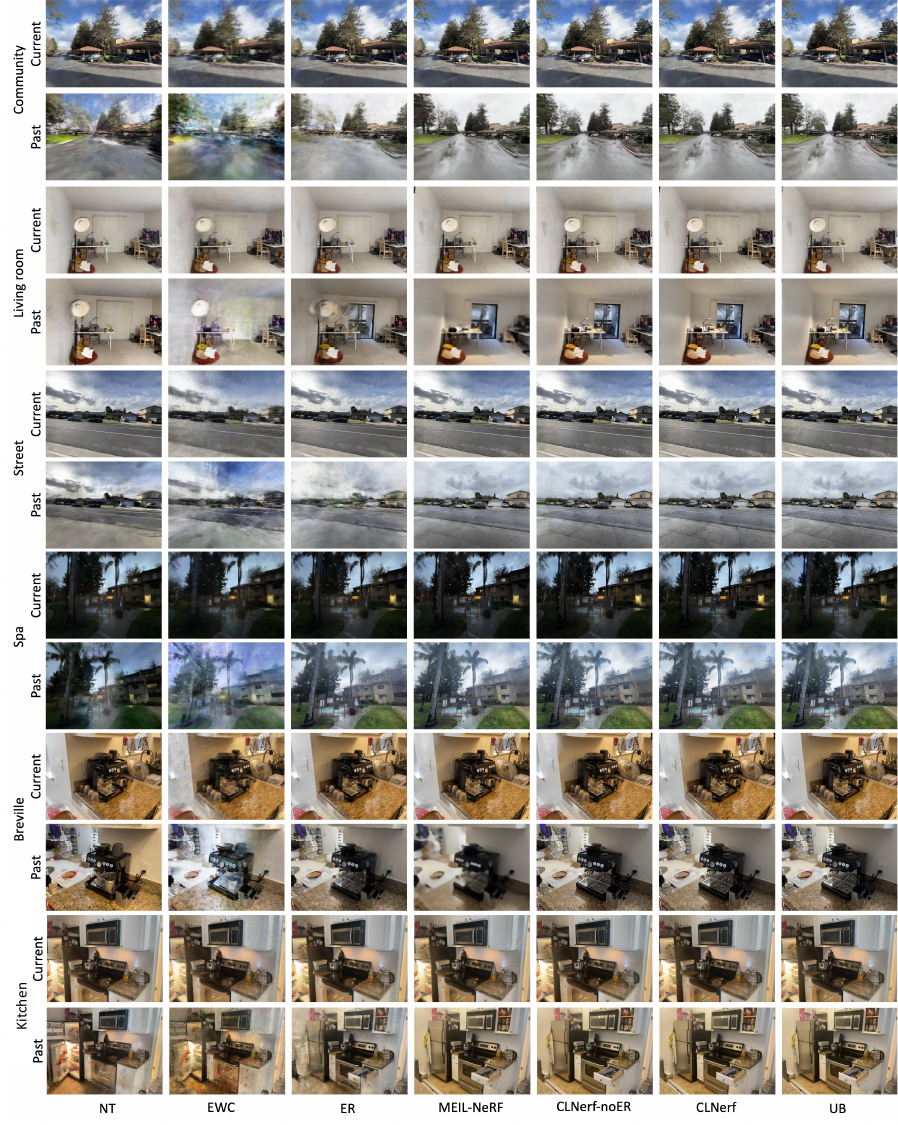}
	\caption{\textbf{Further qualitative results with trainable embeddings used in all methods (zoom-in recommended).} Each two rows show for different methods the rendered views on test data from current and previous steps. Similar performance difference can be observed as in the main experiments.}\label{fig:more_qualitative}
\end{figure*}

Here, we show further qualitative results on individual scenes of WAT. To better demonstrate the advantage of CLNeRF in terms of architecture design and continual learning strategies, we show the results of other continual NeRF methods with (Fig.~\ref{fig:more_qualitative}) and without (Fig.~\ref{fig:more_qualitative_noEmbed}) using the proposed trainable embeddings (please refer to the video demo in our github repo for close-view comparisons). Methods without trainable embeddings cannot recover the geometry and appearance of past time steps, resulting in severe artifacts. Even with trainable embeddings, severe artifacts (NT, EWC) and detail lost (ER, MEIL-NeRF) still exist in other baselines.

\section{Quantitative Results on Individual Scenes} \label{appdx:more_qualitative}

\begin{table*}
\centering
\begin{tabular}{c|cccccc}
\toprule
 \diagbox{Method}{Scene} & Breville & Kitchen & Living room & Community & Spa & Street \\
\midrule
NT (NeRF) noEmbed & 19.53/0.676 & 18.30/0.771 & 17.27/0.788 & 13.58/0.533 & 12.97/0.489 & 15.36/0.490 \\ 
NT (NeRF) & 19.95/0.695 & 18.69/0.771 & 17.65/0.790  & 13.97/0.538 & 14.71/0.612 & 15.24/0.486 \\ 
EWC (NeRF) noEmbed & 19.07/0.653 & 17.70/0.753 & 17.36/0.782  & 13.40/0.530 & 12.57/0.501 & 14.76/0.473 \\ 
EWC (NeRF) & 19.15/0.657 & 17.89/0.745 & 17.47/0.780  & 14.12/0.543 & 17.04/0.701 & 15.14/0.475 \\ 
ER (NeRF) noEmbed & 19.07/0.673 & 20.12/0.793 & 17.84/0.793  & 16.68/0.589 & 16.77/0.634 & 16.87/0.512 \\ 
ER (NeRF) & 23.46/0.717 & 25.03/0.836 & 21.41/0.813  & 17.65/0.600 & 21.92/0.752 & 18.71/0.538 \\ 
MEIL-NeRF (NGP) noEmbed & 22.12/0.781 & 22.08/0.822 & 18.11/0.785 & 21.91/0.612 & 20.67/0.741 & 20.64/0.581 \\ 
MEIL-NeRF (NGP) & 23.55/0.710 & 27.27/0.861 & 23.56/0.821 & 21.92/0.609 & 25.60/0.771 & 22.39/0.608 \\ 
CLNeRF-noER (NGP) & 27.61/0.808 & \textbf{28.59/0.880} & 24.56/0.829 & 22.80/0.628 & \textbf{26.56/0.812} & \textbf{22.53/0.615} \\ 
CLNeRF (NGP) & \textbf{28.02/0.826} & 28.40/0.877 & \textbf{24.58/0.829} & \textbf{22.88/0.629} & 26.28/0.811 & 22.53/0.612 \\  
\midrule
UB (NGP) & 28.62/0.838 & 28.53/0.878 & 24.51/0.826 & 23.65/0.634 & 26.92/0.812 & 22.87/0.615 \\ 
\bottomrule
\end{tabular}
\caption{\textbf{Results on individual scenes of WAT}.}\label{tab:per_scene_WAT}
\end{table*}

\begin{table*}
\centering
\footnotesize{
\begin{tabular}{c|cccccccc}
\toprule
 \diagbox{Method}{Scene} & Lego & Chair & Drums & Ficus & Hotdog & Materials & Mic & Ship \\
\midrule
NT (NeRF) & 30.70/0.956 & 31.02/0.965 & 21.96/0.897 & 27.47/0.957 & 32.64/0.969 & 26.63/0.932 & 29.68/0.972 & 28.17/0.855 \\ 
EWC (NeRF) & 30.82/0.952 & 30.57/0.950 & 22.58/0.895 & 26.41/0.946 & 32.55/0.967 & 26.39/0.918 & 30.73/0.970 & 26.55/0.823 \\ 
ER (NeRF) & 33.21/0.969 & 32.76/0.973 & 23.25/0.915  & 29.56/0.968 & 34.22/0.973 & 27.50/0.939 & 32.79/0.984 & 29.06/0.866 \\ 
MEIL-NeRF (NGP) & 32.91/0.971 & 32.75/0.979 & 24.35/0.928 & 30.96/0.977 & 34.87/0.977 & 28.16/0.941 & 32.93/0.986 & 28.52/0.866 \\ 
CLNeRF-noER (NGP) & 34.34/0.975 & 34.29/\textbf{0.982} & \textbf{25.51/0.931} & 32.91/0.979 & 36.20/0.979 & 28.92/0.941 & 34.38/0.987 & 29.15/0.878 \\ 
CLNeRF (NGP) & \textbf{34.80/0.976} & \textbf{34.39}/0.980 & 25.46/0.931 & \textbf{33.10/0.980} & \textbf{36.42/0.979} & \textbf{29.12/0.943} & \textbf{34.68/0.987} & \textbf{29.30/0.880} \\  
\midrule
UB (NGP) & 35.70/0.978 & 35.41/0.982 & 25.74/0.932 & 33.96/0.982 & 37.24/0.980 & 29.43/0.944 & 35.83/0.989 & 30.19/0.888 \\ 
\bottomrule
\end{tabular}
}
\caption{\textbf{Results on individual scenes of SynthNeRF}.}\label{tab:per_scene_SynthNeRF}
\end{table*}

\begin{table*}
\centering
\begin{tabular}{c|cccccc}
\toprule
 \diagbox{Method}{Scene} & Ignatius & Truck & Barn & Caterpillar & Family \\
\midrule
NT (NeRF) & 24.51/0.924 & 16.76/0.739 & 15.98/0.614 & 13.64/0.701 & 22.03/0.863 \\ 
EWC (NeRF) & 24.83/0.925 & 16.28/0.730 & 12.46/0.619  & 14.34/0.898 & 22.68/0.880 \\ 
ER (NeRF) & 26.79/0.948 & 24.26/0.883 & 26.02/0.836  & 24.75/0.898 & 30.71/0.944 \\ 
MEIL-NeRF (NGP) & 29.56/0.954 & 25.85/0.903 & 22.99/0.825 & 26.33/0.916 & 31.24/0.949 \\ 
CLNeRF-noER (NGP) & 30.22/0.956 & 27.55/\textbf{0.920} & \textbf{27.55/0.851} & 28.12/0.930 & 33.34/0.961 \\ 
CLNeRF (NGP) & \textbf{30.41/0.957} & \textbf{27.59}/0.918 & 27.47/0.848 & \textbf{28.29/0.933} & \textbf{33.68/0.961} \\  
\midrule
UB (NGP) & 30.94/0.960 & 28.11/0.926 & 28.45/0.866 & 29.08/0.940 & 34.91/0.964 \\ 
\bottomrule
\end{tabular}
\caption{\textbf{Results on individual scenes of NSVF}.}\label{tab:per_scene_NSVF}
\end{table*}

\begin{table*}
\centering
\begin{tabular}{c|ccccc}
\toprule
 \diagbox{Method}{Scene} & M60 & Playground & Train & Truck \\
\midrule
NT (NeRF) & 15.87/0.569 & 15.70/0.444 & 12.73/0.365 & 14.91/0.468 \\ 
EWC (NeRF) & 13.89/0.465 & 16.28/0.462 & 13.38/0.366 & 16.56/0.478 \\ 
ER (NeRF) & 16.10/0.580 & 19.67/0.569 & 16.18/0.476 & 17.46/0.544 \\ 
MEIL-NeRF (NGP) & 18.11/0.621 & 21.53/0.592 & 17.16/0.533 & 20.76/0.635 \\ 
CLNeRF-noER (NGP) & 18.88/0.631 & 22.18/0.643 & 17.20/0.561 & \textbf{22.99}/0.694 \\ 
CLNeRF (NGP) & \textbf{19.04/0.634} & \textbf{22.37/0.643} & \textbf{17.31/0.563} & 22.61/\textbf{0.695} \\
\midrule
UB (NGP) & 18.69/0.623 & 22.31/0.672 & 17.36/0.586 & 22.99/0.712 \\ 
\bottomrule
\end{tabular}
\caption{\textbf{Results on individual scenes of NeRF++}.}\label{tab:per_scene_NeRF++}
\end{table*}

\begin{table*}
\centering
\begin{tabular}{c|ccccc}
\toprule
 \diagbox{Method}{Scene} & Brandenburg Gate & Sacre Coeur & Trevi Fountain & Taj Mahal \\
\midrule
NT (NGP) & 21.11/0.793 & 15.78/0.642 & 19.43/0.613 & 19.82/0.722 \\ 
ER (NGP) & 24.20/0.803 & 16.91/0.682 & 19.57/0.616 & 19.85/0.723 \\ 
MEIL-NeRF (NGP) & 24.22/0.802 & 20.53/0.744 & 21.42/0.667 & 23.21/0.770 \\ 
CLNerf-noER (NGP) & 25.24/\textbf{0.803} & 20.62/0.753 & \textbf{21.53/0.668} & 23.32/\textbf{0.781} \\ 
CLNerf (NGP) & \textbf{25.43}/0.802 & \textbf{21.32/0.765} & 21.44/0.667 & \textbf{23.34}/0.780 \\  
\midrule
UB (NeRFW) & 24.23/0.881 & 21.59/0.833 & 21.99/0.853 & 23.32/0.726 \\ 
UB (NGP) & 25.58/0.813 & 21.22/0.785 & 21.64/0.676 & 23.76/0.780 \\ 
\bottomrule
\end{tabular}
\caption{\textbf{Results on individual scenes of Phototourism}.}\label{tab:per_scene_phototour}
\end{table*}

This section shows the quantitative results (Tab.~\ref{tab:per_scene_WAT} to~\ref{tab:per_scene_phototour}) on individual scenes of each used dataset. CLNeRF performs better than other continual NeRF approaches on \emph{all} individual scenes, even without storing any historical image. The performance gap between CLNeRF and UB is also small for all scenes. Due to the training noise and the close practical performance, the best model of each scene changes between CLNeRF and CLNeRF-noER and UB.

\section{Baseline Results with NGP} \label{appdx:ngp_baseline}

In the main experiments (Tab.~\ref{tab:main}), we choose the best architecture for different baselines and observe that methods like ER and NT perform better with vanilla NeRF. Here, we further report the baseline results using NGP architecture. Due to the NaN loss issue of NGP, we cannot perform EWC on it. Hence, we only report the results for NT and ER. Due to the resource limit, we only report the results on Synth-NeRF, NeRF++ and WAT. As discussed in \cref{sec:exp_arch}, NGP without generative replay overfits to the training views and fails to generalize to novel views and past time steps.

\begin{table}[t]
\centering
\small{
\begin{tabular}{c|ccc}
\toprule
 Dataset & Synth-NeRF & NeRF++ & WAT \\
\midrule
NT (NGP) & 21.66/0.858 &  11.93/0.380 & 11.51/0.356 \\ 
ER (NGP) & 27.35/0.919 & 15.13/0.436 & 19.03/0.657 \\ 
\bottomrule
\end{tabular}
}
\vspace{3pt}
\caption{\textbf{Baseline results on NGP architecture}. Comparing with Tab.~\ref{tab:main}, we can see that for baselines like NT and ER, NGP performs much worse than vanilla NeRF. See Sec.~\ref{sec:exp_arch} for analysis on architectures.}\label{tab:ngp_baseline}
\vspace{-1em}
\end{table}

\section{More WAT Scenes}

In this section, we add 4 more scenes to WAT, making the total number of scenes from 6 to 10. We use the same data creation process as described in~\cref{sec:CLRender}, and call this enlarged dataset WAT+ (available in our code repository). As shown in Tab.~\ref{tab:WAT+}, the results on these new scenes are consistent with Tab.~\ref{tab:main}.

\begin{table}
\centering
\scriptsize{
\begin{tabular}{c|cccc}
\toprule
 \diagbox{Method}{Scene} & Car & Grill & Mac & Ninja \\
\midrule
NT (NeRF) & 19.14/0.516 & 19.96/0.612 & 18.68/0.831 & 19.92/0.814 \\ 
EWC (NeRF) & 18.46/0.500 & 19.67/0.602 & 18.48/0.816 & 19.74/0.805 \\ 
ER (NeRF) & 19.10/0.510 & 21.03/0.618 & 23.07/0.871 & 21.49/0.827 \\ 
MEIL-NeRF (NGP) & 21.80/0.528 & 24.01/0.648 & 12.71/0.687 & 13.22/0.668 \\ 
CLNeRF-noER (NGP) & 22.63/0.539 & 24.81/0.652 & \textbf{29.34/0.907} & 26.42/0.869 \\ 
CLNeRF (NGP) & \textbf{22.73/0.541} & \textbf{24.84/0.653} & 29.33/0.906 & \textbf{27.19/0.878} \\  
\midrule
UB (NGP) & 22.62/0.538 & 25.12/0.661 & 30.16/0.905 & 26.71/0.875 \\ 
\bottomrule
\end{tabular}
}
\caption{\textbf{Results on the 4 scenes added to WAT+}.}\label{tab:WAT+}
\end{table}

\end{document}